\documentclass[sigconf]{acmart}

\renewcommand\footnotetextcopyrightpermission[1]{}
\AtBeginDocument{%
  }

\usepackage{multirow}
\usepackage{booktabs}
\usepackage{balance}
\usepackage{makecell}

\usepackage{booktabs}
\usepackage{makecell}
\usepackage{graphicx}
\usepackage{algorithm}
\usepackage[noend]{algpseudocode}
\usepackage{enumitem}
\usepackage{amsthm}
\newtheorem{definition}{Definition}

\newcommand{\ie}{i.e.\xspace}
\newcommand{\eg}{e.g.\xspace}
\newcommand{\ourmethod}{TRACE\xspace}

\usepackage{xspace}
\usepackage{multirow}
\usepackage{subfig}
\usepackage{caption}

\usepackage{caption}
\usepackage{enumitem}
\setlist[itemize]{nosep, topsep=2pt, partopsep=0pt, parsep=0pt, itemsep=2pt}

\begin{document}

\acmConference[]{}{}{}

\title{
TRACE: Learning to Compute on Circuit Graphs}


\author{Ziyang Zheng\textsuperscript{*} \quad Jiaying Zhu\textsuperscript{*}\quad Jingyi Zhou \quad Qiang Xu }
\affiliation{%
  \institution{The Chinese University of Hong Kong}
  \country{}
}

\begin{abstract}

Learning to compute, the ability to model the functional behavior of a circuit graph, is a fundamental challenge for graph representation learning. Yet, the dominant paradigm is architecturally mismatched for this task. This flawed assumption, central to mainstream message passing neural networks (MPNNs) and their conventional Transformer-based counterparts, prevents models from capturing the position-aware, hierarchical nature of computation.
To resolve this, we introduce \textbf{TRACE}, a new paradigm built on an architecturally sound backbone and a principled learning objective. First, TRACE employs a Hierarchical Transformer that mirrors the step-by-step flow of computation, providing a faithful architectural backbone that replaces the flawed permutation-invariant aggregation. Second, we introduce \textbf{function shift learning}, a novel objective that decouples the learning problem. Instead of predicting the complex global function directly, our model is trained to predict only the \textit{function shift}, the discrepancy between the true global function and a simple local approximation that assumes input independence. We validate this paradigm on various circuits modalities, including Register Transfer Level graphs, And-Inverter Graphs and post-mapping netlists. Across a comprehensive suite of benchmarks, TRACE substantially outperforms all prior architectures. These results demonstrate that our architecturally-aligned backbone and decoupled learning objective form a more robust paradigm for the fundamental challenge of learning the functional behavior of a circuit graph. Code is available at \href{https://github.com/zyzheng17/TRACE_DAC26}{\textit{\textcolor{blue}{https://github.com/zyzheng17/TRACE\_DAC26}}}.

\end{abstract}

\maketitle

\makeatletter
\def\blfootnote{\xdef\@thefnmark{}\@footnotetext}
\makeatother
\blfootnote{$^{*}$Equal contribution.}
\section{Introduction}
\label{sec:intro}

Circuit graphs provide a fundamental abstraction for modeling computation. As directed graphs of nodes representing operations and variables, they capture the flow of computation and are crucial in various circuit modalities. Accurately modeling the computational behavior of circuit graphs is therefore a critical enabler for high-impact applications, including performance prediction~\cite{chen2024large, xie2022placement, zhang2020grannite, mendis2019ithemal}, verification~\cite{li2023deepsat,Selsam2018LearningAS,zhang2021simulate}, and optimization~\cite{zuo2023mul,Mirhoseini2021AGP}.

This need for functional modeling has driven recent work in graph representation learning, with approaches largely divided into two families: message passing neural networks (MPNNs) and Graph Transformers. These models have been increasingly applied to capture the functionality of diverse computational graph modalities, with a significant body of work focusing on the particularly challenging domain of hardware design, including Register Transfer Level (RTL) graphs~\cite{fang2025circuitencoder,fang2025circuitfusion}, And-Inverter Graphs (AIGs)~\cite{shi2023deepgate2,shi2024deepgate3,zheng2025deepgate4,khan2025deepseq2,wu2025MGVGA,wang2024fgnn2,PolarGate}, and post-mapping (PM) netlists~\cite{shi2025deepcell}. 
MPNNs provide a general framework~\cite{gilmer2017neural} based on a paradigm of aggregating and updating node features:
\begin{equation}
    \mathbf{x}_i' = \gamma \left( \mathbf{x}_i, \, \underset{j \in \mathcal{N}(i)}{\square} \, \varphi \left( \mathbf{x}_i, \mathbf{x}_j, \mathbf{e}_{j,i} \right) \right),
\end{equation}
where $\square$ denotes a permutation-invariant aggregator (\eg, sum, mean, max) and the update function $\gamma$ and message function $\varphi$ are differentiable functions. In contrast, Transformer-based methods either flatten the graph into a sequence for global self-attention~\cite{fang2025circuitencoder,fang2025circuitfusion}, or incorporate graph structure via attention masks~\cite{shi2024deepgate3,zheng2025deepgate4,fang2025nettag}.

\begin{figure}
    \centering
    \includegraphics[width=\linewidth]{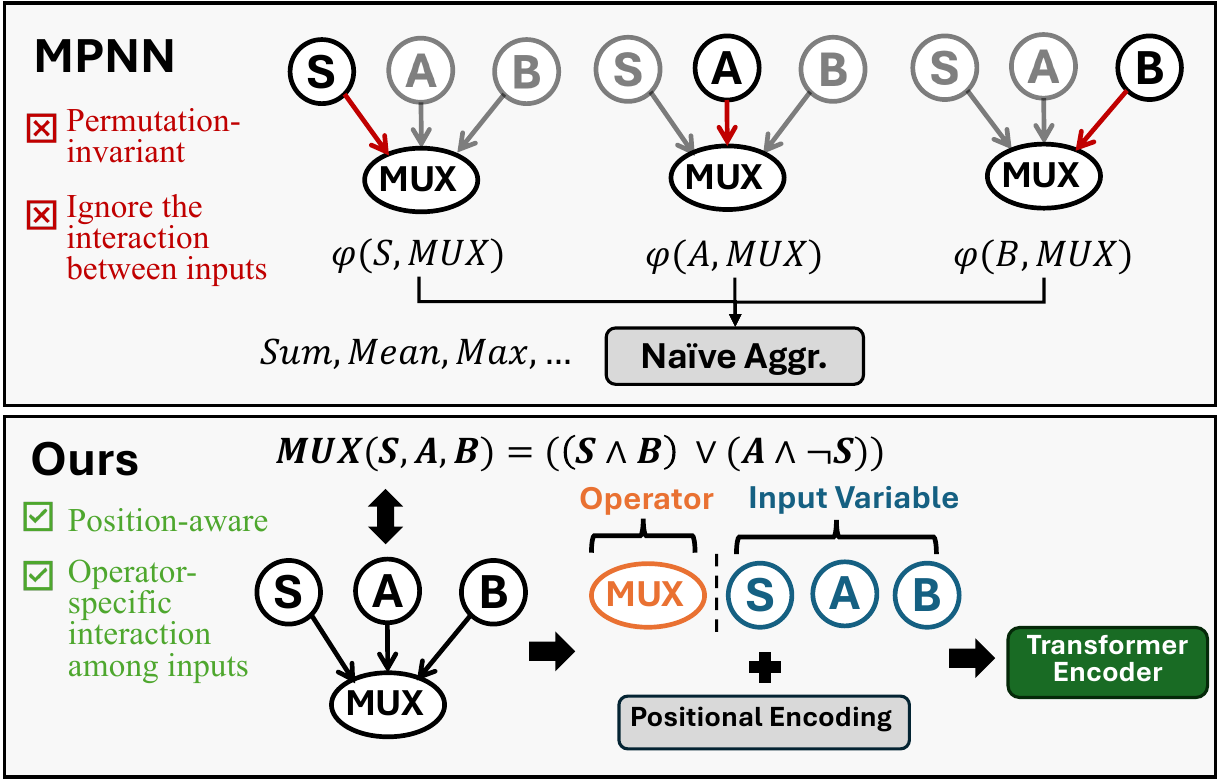}

    \caption{The architectural failure of MPNNs on circuit graphs. \textbf{Top:} The permutation-invariant aggregation in MPNNs cannot distinguish between ordered inputs (e.g., $\texttt{A,B}$ vs $\texttt{B,A}$), yielding the same incorrect embedding for a position-aware operator like MUX. \textbf{Bottom:} Our approach processes inputs as an ordered sequence, enabling position-awareness and capturing operator-specific interactions.}
    \label{fig:ours_mpnn}

\end{figure}

Despite their success on general graph domains, we argue that these architectures are ill-suited for computation due to a \textbf{fundamental mismatch} between their properties and the nature of computation process in circuit graphs (as shown in Fig.~\ref{fig:ours_mpnn}):

\begin{enumerate}
   
    \item \textbf{MPNNs Fail to Model Input Interactions:} The core architectural flaw of MPNNs is that their message functions model interactions between an operator and only a single input at a time. These independent messages are then combined by a permutation-invariant aggregator. This two-stage process makes it architecturally impossible to capture the intricate, operator-specific relationships between multiple inputs and, as a direct consequence, to model position-aware operators like the MUX (multiplexer), where input order is critical (i.e., $\texttt{MUX(S, A, B)} \neq \texttt{MUX(A, S, B)}$).
    
    \item \textbf{Vanilla Transformers are Hierarchy-Agnostic:} Fully-connected Transformers flatten the graph into a sequence, destroying the explicit hierarchy and connectivity for modeling computation. For an expression like $y = f(g(a,b), c)$, this architecture fails to capture the intermediate dependency on $g(a,b)$ and cannot guarantee a correct computational trace.
    
    \item \textbf{Edge-masked Transformers Inherit MPNN Limitations:} Edge-masked transformers, while respecting connectivity, often reduce to GAT-like attention mechanisms. This is functionally a weighted sum, which still inherits the fundamental limitations of the message-passing paradigm, failing to model the precise, non-linear interactions required by logical and algebraic operators.
\end{enumerate}

These architectural flaws highlight a failure to model the step-by-step flow of computation. Beyond this, however, a deeper challenge lies in capturing global function that emerges from the graph's overall topology. Even with simple components (\eg, \textsc{AND} and \textsc{NOT} gates in AIGs), a graph's overall functionality can become highly complex due to reconvergent dependencies.
For instance, consider $c = a \land b$ with $a = x \land y$ and $b = y \land z$, where $x,y,z \sim \mathcal{B}(p)$ (a Bernoulli distribution with parameter $p$). Locally, $a,b \sim \mathcal{B}(p^2)$, and ignoring reconvergence, one would predict $c \sim \mathcal{B}(p^4)$. However, since $y$ appears in both $a$ and $b$, $a$ and $b$ are correlated, shifting the true distribution to $c \sim \mathcal{B}(p^3)$.
This \textit{function shift}, from $\mathcal{B}(p^4)$ to $\mathcal{B}(p^3)$, demonstrates how dependencies fundamentally alter functional behavior. Previous works that supervise directly on a node's final global function implicitly bundle this effect into the embeddings, making it difficult for the model to distinguish true functional dependencies from spurious correlations.

Our work addresses these challenges with \textbf{\ourmethod}, a \underline{T}ransformer for \underline{R}easoning about \underline{A}lgebraic and \underline{C}omputational \underline{E}xpressions, which presents a two-fold solution. First, to resolve local architectural flaws of prior models, \ourmethod~employs a hierarchical Transformer. Inspired by prefix notation, we represent a computation step as an ordered sequence, $\texttt{[operator, input\_1, input\_2, \dots]}$, processed by a Transformer encoder with positional encoding (Figure~\ref{fig:ours_mpnn}). Applying this process recursively according to graph's logical dependencies, \ourmethod~learns a faithful, position-aware representation of each computational step. Second, to capture the global function, we introduce \textbf{function shift learning}, a novel objective that explicitly models the discrepancy between local and global functions, allowing the model to disentangle a node's intrinsic behavior from the contextual effects imposed by wider graph topology.

Our experiments span a broad spectrum of circuit modalities (RTL, AIG, and PM netlists) and tasks (contrastive and predictive) across several standard benchmarks, including ITC~\cite{corno2002ITC}, OpenCores~\cite{albrecht2005opencore}, ISCAS '89~\cite{brglez1989ISCAS}, ForgeEDA~\cite{shi2025forgeeda}, and DeepCircuitX~\cite{li2025deepcircuitx}. 
The results are unequivocal: 
\ourmethod~consistently and substantially outperforms prior approaches across all settings. This establishes \ourmethod~not only as a new state of the art for circuit analysis, but as a more robust and architecturally sound paradigm for learning on circuit graphs.
\section{Background}
\label{sec:background}
\begin{figure}[h]

    \centering
    \includegraphics[width=\linewidth]{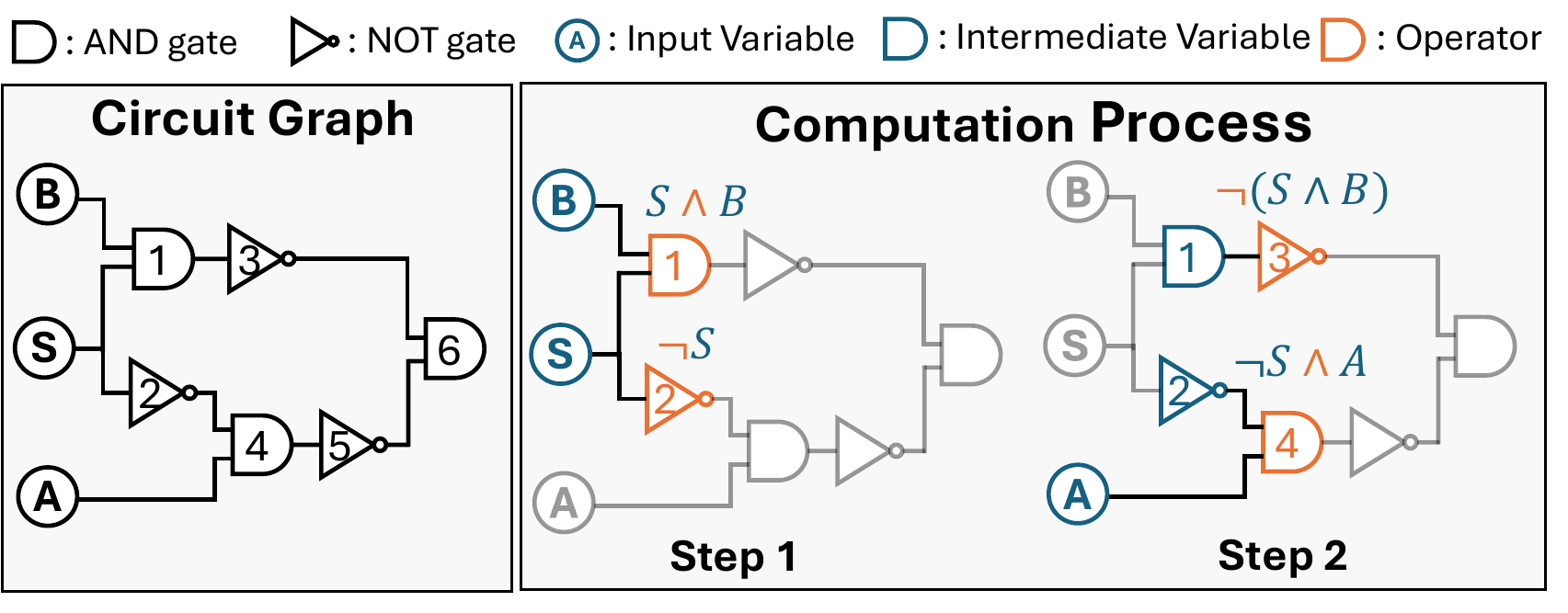}
    \caption{Illustration of a circuit graph and its computation process.
    This figure demonstrates the dual role of nodes within a circuit graph.
    \textbf{Step 1} shows nodes 1 and 2 functioning as operators to compute the expressions $S \land B$ and $\lnot S$, respectively.
    As computation progresses to \textbf{Step 2}, these nodes transition to representing the intermediate variables that hold the results of these operations in step 1, which are then passed to subsequent operators for further computation. }
    \label{fig:computational_graph}
    \vspace{-6pt}
\end{figure}

\subsection{Circuit Graphs}
In our work, circuit graphs are defined as directed graphs where nodes represent either input variables or operators, and edges signify data flow. As depicted in Figure~\ref{fig:computational_graph}, this structure allows a single node to serve dual roles: it can be an operator receiving data from its source nodes and an intermediate variable whose output is consumed by other operators. Specifically, for any directed edge from a source to a target node, the target always represents an operator, while the source represents a variable—either an initial input variable or a temporary intermediate variable resulting from a previous operation. This representation effectively models the data dependencies and computational flow within the system. We focus on three types of circuit graphs from the front-end of the electronic design automation (EDA) flow: Register-Transfer Level (RTL) graphs, And-Inverter Graphs (AIGs), and Post-Mapping (PM) netlists. 


\subsection{Message Passing Neural Networks}
Message Passing Neural Networks (MPNNs) are a dominant architectural paradigm for function learning on circuit graphs. These models can be broadly categorized into two types: synchronous and asynchronous. Synchronous MPNNs~\cite{wu2023gamora, PolarGate, wu2025MGVGA, hoga} process all message-passing updates in parallel, a strategy designed for computational efficiency. In contrast, asynchronous MPNNs~\cite{li2022deepgate,shi2023deepgate2, shi2025deepcell,khan2025deepseq2,wang2024fgnn2} mimic the logic simulation process by updating node representations sequentially, following a topological order, which aims to capture effective functional representations by emulating the data flow. 
Such asynchronous processing may introduces a potential limitation for processing circuits with significant logic depth. This limitation is often mitigated in practice, as real-world timing constraints typically restrict the maximum number of logic levels.
Despite their architectural differences, both synchronous and asynchronous MPNNs rely on the conventional message passing paradigm~\cite{gilmer2017neural} (Figure~\ref{fig:ours_mpnn}). This paradigm fundamentally struggles to capture the operator-specific and position-aware interactions among input variables.

\begin{figure*}[h]
    \centering
    \includegraphics[width=0.95\linewidth]{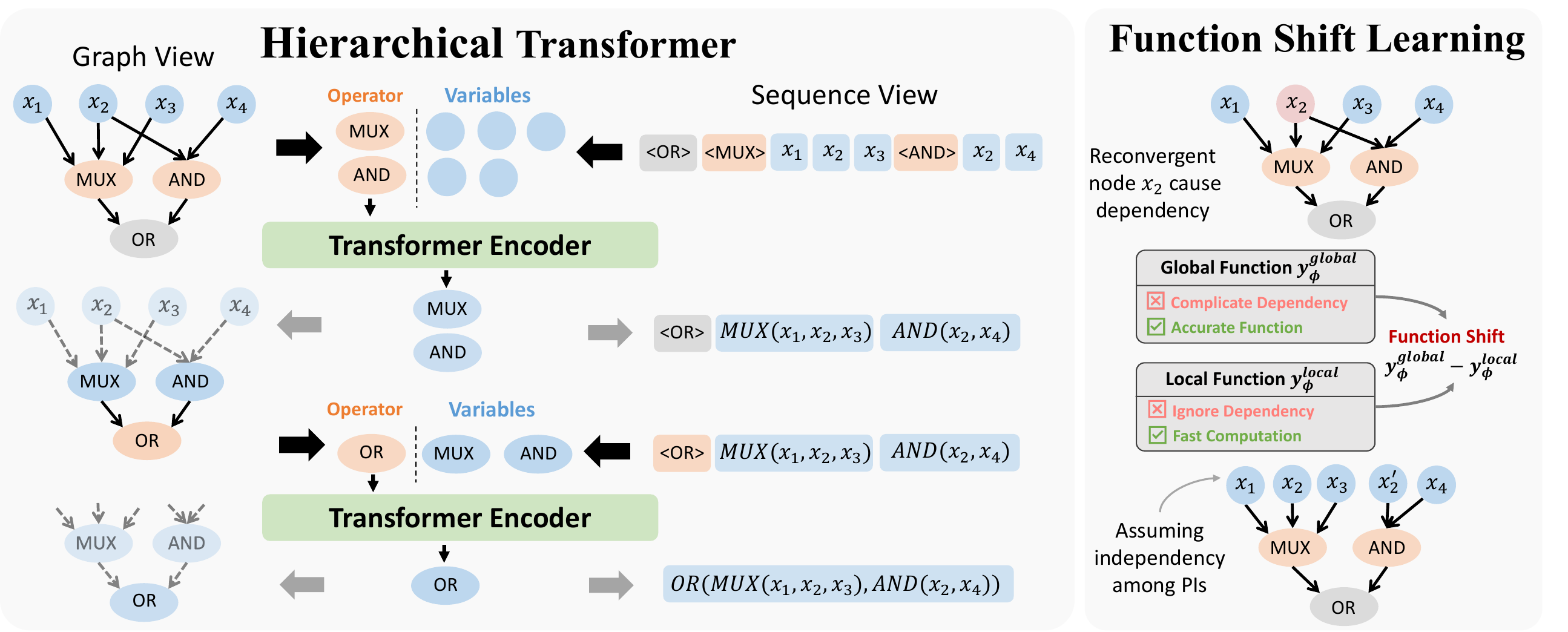}
    \caption{%
    Overview of our proposed framework. 
    \textbf{Left:} A circuit graph, represented in both a graph view and its equivalent prefix notation, is encoded by a Hierarchical Transformer to model the computation process. 
    \textbf{Right:} For predictive tasks, we introduce Function Shift Learning (FSL). Instead of directly regressing the global function, the model captures the difference between the global and local functions: $y^{FSL}=y^{global}-y^{local}$. 
     }
    \label{fig:framework}
    \vspace{-2pt}
\end{figure*}

\subsection{Graph Transformers}

Despite the widespread use of MPNNs, they suffer from inherent limitations, including difficulty in capturing long-range dependencies and susceptibility to over-smoothing~\cite{li2018deeper} and over-squashing~\cite{alon2020bottleneck}. These drawbacks have motivated a shift towards Graph Transformers, which leverage global attention to address them. Existing graph transformers fall into two categories: fully-connected transformers and edge-masked transformers.

Fully-connected transformers, such as \citet{rampavsek2022graphgps,wu2023sgformer, wu2023difformer} for general graphs and \citet{fang2025circuitencoder,fang2025circuitfusion} for circuit graphs, treat the graph as a flattened sequence of nodes. While this enables global self-attention, it inadvertently leads to a loss of the critical hierarchical structure and intermediate dependencies that are inherent to circuit graphs. In contrast, edge-masked transformers, including DeepGate3~\cite{shi2024deepgate3}, DeepGate4~\cite{zheng2025deepgate4}, and NetTAG~\cite{fang2025nettag}, integrate the graph's topology by using the adjacency matrix to mask the attention mechanism. However, this approach often reduces the model to a Graph Attention Network (GAT)-like attention mechanism, inheriting the limitations of traditional MPNNs.
\section{Method}




\subsection{Hierarchical Transformer}
\label{sec:htf}

To address the limitations of MPNNs and Graph Transformers discussed in Section~\ref{sec:intro} and ~\ref{sec:background}, we propose a new paradigm for encoding circuit graphs, with a position-aware Hierarchical Transformer that enables operator-specific interaction among input variables,  faithfully mirroring the computational process illustrated in Figure~\ref{fig:computational_graph}.

A circuit graph $\mathcal{G}=(\mathbf{V},\mathbf{E})$, consists of primary input (PI) nodes, which have an in-degree of zero, and operator nodes, whose in-degree is determined by their type. We begin by computing the logic level of each node in topological order\footnote{To handle cyclic graphs, such as sequential AIGs, we follow \citet{khan2025deepseq2} by treating all flip-flops or registers as pseudo Primary Inputs (PIs) and removing the feedback edges to compute the logical level.} as follows:
\vspace{-2pt}
\begin{equation}
level(v) =
\begin{cases} 
0 & \text{if $v$ is PI} \\
1 + \max\limits_{(u,v) \in \mathbf{E}} level(u) & \text{otherwise}
\end{cases}
\end{equation}

The logic level defines the dataflow through the graph. Inspired by asynchronous MPNNs, we process nodes level by level. First, Primary Input (PI) nodes (level 0) are initialized with their input distribution, e.g., $\mathcal{B}(p)$, a Bernoulli distribution with parameter $p$. Then, for the nodes $v_0^k, v_1^k, \dots, v_{n_k}^k$ at subsequent level $k>0$, we gather their respective sets of direct predecessors, $\mathcal{N}(v_0^k), \dots, \mathcal{N}(v_{n_k}^k)$, where each set is defined as $\mathcal{N}(v_i^k) = \{u_j \in \mathbf{V} \mid (u_j,v_i^k)\in \mathbf{E} \}$.  

Inspired by prefix notation, we represent each computation at level $k$ as an ordered sequence, $\mathbf{v}_i^k$, formed by placing the \emph{operator} $v_i^k$ at the head, followed by its \emph{ordered input nodes} from $\mathcal{N}(v_i^k)$:
\begin{equation}
    \mathbf{v}_i^k = [v_i^k, u_{j_1}, u_{j_2}, \dots, u_{j_{|\mathcal{N}(v_i^k)|}}], \text{ where } u_{j_1},\dots,u_{j_{|\mathcal{N}(v_i^k)|}} \in \mathcal{N}(v_i^k)
\end{equation}

To model operator-specific interactions among inputs, we apply a Transformer encoder to the sequence $\mathbf{v}_i^k$ augmented with positional encodings. The updated embedding for the operator node $v_i^k$ is taken from the Transformer's output corresponding to the first token:
\vspace{-3pt}
\begin{equation}
    v_i^k = \text{Transformer}(\mathbf{v}_i^k + \mathbf{pos})[0],
\end{equation}
where $\mathbf{pos}$ represents the positional encodings. The input sequence $\mathbf{v}_i^k$ is composed of the initial embedding of the operator $v_i^k$ (e.g., a one-hot vector of its type) and the embeddings of its input nodes $\{u_{j_m}\}$, which have been computed in previous steps. After updating, the embedding of $v_i^k$ now represents the result of the computation at this node, \ie an intermediate variable node.

This paradigm offers several key advantages. First, positional encodings enable position-aware aggregation for order-dependent operations, contrasting with the permutation-invariant nature of standard message passing. Second, self-attention enables rich interactions among all source nodes [$u_{j_1}, \dots, u_{j_{|\mathcal{N}(v_i^k)|}}$], faithfully mirroring the computational dataflow of an operator. Finally, the sequence length depends only on in-degree: unlike general graphs with long-tailed degree distributions that incur high overhead, circuit graphs have small, tightly bounded in-degrees, preventing the memory explosion caused by quadratic complexity.

\vspace{2pt}
\textit{\textbf{An Alternative View: Hierarchical Transformer on Prefix Notation.}}
Our proposed method can also be interpreted as a Hierarchical Transformer operating on prefix notation, in a way that follows the logical dependencies inherent in the circuit graph. This perspective highlights how the model processes the graph by mirroring the step-by-step evaluation of an expression. A circuit graph can be converted into a prefix notation string through a pre-order traversal on its reversed edges. As illustrated in Figure~\ref{fig:framework}, our proposed Hierarchical Transformer, unlike methods that simply flatten the graph into a single sequence, inherently preserves the nested, hierarchical structure of dependencies within each computation step, which allows it to align naturally with the actual flow of computation. This view also underscores the generalizability of our approach to other sequence-based problems, such as hardware model-checking, where the problem is often represented in Btor2 format~\cite{ArminBiere2018Btor2}, a prefix-style representation of bit-vector formulas.

\subsection{Function Shift Learning on Predictive Task} 
\label{sec:FSL}

Logic-1 probability prediction is a task widely studied in prior works~\cite{shi2023deepgate2,khan2025deepseq2,shi2024deepgate3,zheng2025deepgate4,shi2025deepcell,PolarGate}, as it serves as a key indicator of a model’s ability to capture circuit functionality. 
The logic-1 probability corresponds to the \emph{global function} of a circuit, as defined in Definition~\ref{def:global_func}. This function can be highly complex due to reconvergent dependencies, and computing it directly requires enumerating the joint distribution of all inputs, which incurs an exponential cost of $O(2^k)$.

\begin{definition}[\textbf{Global Function}]
    Given an operator $\phi$, input variables $\mathbf{x}=[x_1,x_2,\dots,x_k]$ and its distribution $\mathcal{D}$, the global function is defined as
    $y_{\phi}^{global} = \mathbb{E}_{\mathbf{x}\sim \mathcal{D}}[\phi(x_1,x_2,\dots,x_k)].$
    
\label{def:global_func}
\end{definition}

By ignoring the dependencies among input, \ie by assuming they are independent, we can derive the \emph{local function}, which is formally stated in Definition~\ref{def:local_func}. Although this approximation is computationally efficient with an $O(1)$ complexity, it fails to capture the true function of the circuit.

\begin{definition}[\textbf{Local Function}]
    Given an operator $\phi$, input variables $\mathbf{x}=[x_1,x_2,\dots,x_k]$ and its distribution $\mathcal{D}$, the local function is defined as
    $y_{\phi}^{local} = \phi\!\left(\mathbb{E}_{\mathbf{x}\sim \mathcal{D}}[x_1],\mathbb{E}_{\mathbf{x}\sim \mathcal{D}}[x_2],\dots,\mathbb{E}_{\mathbf{x}\sim \mathcal{D}}[x_k]\right).$
\label{def:local_func}
\end{definition}

Building on these properties, we propose to learn the \emph{function shift}:
$y_{\phi}^{FSL} = y_{\phi}^{global} - y_{\phi}^{local},$
which measures the discrepancy between the local function (Definition~\ref{def:local_func}) and the true global function (Definition~\ref{def:global_func}).
This formulation decouples the global function into two components: a simple local function and a function shift. Rather than predicting the complex global function directly, our model is trained to predict only the function shift. This isolates the complex contextual effects caused by reconvergence, allowing the global function to be reconstructed by simply combining the predicted shift with the local function.

\paragraph{\textbf{Training Stage.}}
During the training stage, the ground-truth global and local functions for each node can be pre-computed from the training data, allowing us to determine the true function shift, $y_i^{FSL}$. We then train the model to regress this value, optimized with an $\mathcal{L}_1$ objective:
\begin{equation}
    \min_\theta \ \mathbb{E}_{\mathcal{G}\sim \mathcal{D}}\Big[ \big|\psi(x_i) - y_i^{FSL}\big| \Big],
\end{equation}
where $x_i$ is the embedding for node $i$ and $\psi(\cdot)$ is a regression head.
\begin{algorithm}
\centering
\caption{Inference with Function Shift}
\label{algo:inference}
\begin{algorithmic}[1]
    \Require Circuit graph $\mathcal{G}=(V,E)$
    \State $\mathbf{x} \gets Hierarchical\_ Transformer (\mathcal{G})$  
    \State $\hat{\mathbf{y}}^{FSL} \gets \psi(\mathbf{x})$
    \State $L \gets \max_{v\in V} \text{level}(v)$
    \For{$l = 1$ to $L$}
        \For{$v \in \{u\in V: \text{level}(u)=l\}$}
            \State 
            $ \hat{y}_{v}^{global} \gets \hat{y}_v^{FSL} 
                + \underbrace{\phi_v(\hat{y}_{u_1}^{global}, 
                \dots, \hat{y}_{u_{|\mathcal{N}(v)|}}^{global})}_{\text{Local Function}}$
        \EndFor
    \EndFor
    \State \textbf{Return} $\hat{\mathbf{y}}^{global}$ for all nodes
\end{algorithmic}

\end{algorithm}


\paragraph{\textbf{Inference Stage.}}
At the inference stage, the true global functions are unknown. Since computing the global function of a node at any given level depends on the global functions of its predecessors from previous levels, we cannot predict them all at once. Therefore, we reconstruct the global functions iteratively, proceeding level by level through the circuit. For each node, we first compute its local function using the already-estimated global functions of its inputs. The final estimate for the node's global function is then obtained by adding the model's predicted function shift to this computed local function. This entire iterative process is detailed in Algorithm~\ref{algo:inference}.

\subsection{Contrastive Task}

Contrastive learning is a standard self-supervised strategy for learning representations of circuit functionality~\cite{wang2024fgnn2,fang2025circuitfusion,fang2025circuitencoder,wu2025MGVGA}. The fundamental principle is to pull embeddings of functionally equivalent circuits closer together in the embedding space while pushing apart those of functionally different circuits. This process trains the model to identify and encode the discriminative features that define a circuit's intrinsic properties, all without needing explicit labels.
Following prior work, we form training instances for each circuit $\mathcal{G}$. A positive sample, $\mathcal{G}^+$, is created by applying a functionally equivalent transformation to $\mathcal{G}$. All other circuits within the same batch are treated as negative samples, $\mathcal{G}^-$. We then optimize the encoder using the InfoNCE loss~\cite{infonce}:
\begin{equation}
    \min_\theta \ \mathbb{E}_{\mathcal{G}\sim \mathcal{D}} \;
    \mathcal{L}_{\text{InfoNCE}}(\mathcal{G}, \mathcal{G}^+, \mathcal{G}^-).
\end{equation}

The objective is to structure the embedding space so that functional equivalence corresponds to proximity, mapping similar circuits to nearby points while separating from non-equivalent ones.
\vspace{-5pt}
\begin{table*}[h]
\setlength{\tabcolsep}{5pt}
\centering
\caption{Comparison of contrastive task across various modalities(\%).}
\label{tab:contrastive_result}
\renewcommand{\arraystretch}{0.85}
\begin{tabular}{lccccccccc}
\toprule
\multicolumn{1}{c}{\multirow{2}{*}{\textbf{Model}}} & \multicolumn{3}{c}{\textbf{RTL}} & \multicolumn{3}{c}{\textbf{AIG}} & \multicolumn{3}{c}{\textbf{Netlist}} \\ \cmidrule(r){2-4} \cmidrule(lr){5-7} \cmidrule(l){8-10}
\multicolumn{1}{c}{} & Rec@1 & Rec@5 & Rec@10 & Rec@1 & Rec@5 & Rec@10 & Rec@1 & Rec@5 & Rec@10 \\ \midrule
\multicolumn{10}{l}{\textit{Message Passing Neural Network}} \\ \midrule
GCN & 82.90 & 87.43 & 91.57 & 83.01 & 93.28 & 96.05 & 58.03 & 77.65 & 85.24 \\
GraphSAGE & 86.46 & 92.87 & 95.43 & 88.55 & 96.41 & 98.38 & 86.12 & 95.82 & 98.13 \\
GAT & 84.98 & 89.22 & 94.65 & 85.68 & 94.32 & 97.60 & 65.21 & 83.72 & 89.97 \\
GIN & 86.23 & 91.34 & 96.95 & 85.98 & 93.40 & 96.52 & 75.85 & 91.82 & 95.90 \\
FGNN2 & - & - & - & 88.73 & 97.03 & 98.57 & - & - & - \\
DeepCell & - & - & - & - & - & - & 80.99 & 95.31 & 97.61 \\ \midrule
\multicolumn{10}{l}{\textit{Graph Transformer}} \\ \midrule
GraphGPS & 86.94 & 92.13 & 96.37 & OOM & OOM & OOM & 45.63 & 61.55 & 69.57 \\
SGFormer & 79.45 & 86.57 & 89.50 & 15.43 & 30.88 & 42.19 & 15.83 & 37.06 & 49.73 \\
DIFFormer & 88.28 & 92.97 & 96.88 & 37.03 & 68.87 & 80.66 & 25.23 & 45.52 & 55.67 \\
CircuitEncoder & 88.27 & 92.97 & 94.52 & - & - & - & - & - & - \\ \midrule
\textbf{\ourmethod} & \textbf{94.45} & \textbf{98.74} & \textbf{99.89} & \textbf{92.68} & \textbf{98.65} & \textbf{99.51} & \textbf{90.81} & \textbf{98.48} & \textbf{99.44} \\ \bottomrule
\end{tabular}
\vspace{-5pt}
\end{table*}

\section{Experiment}

\subsection{Implement Details}
\paragraph{\textbf{Dataset}} In this paper, we conduct experiments on three modalities: RTL, AIG and PM netlist. For RTL, we follow prior works~\cite{fang2025circuitfusion,fang2025circuitencoder} and collect data from ITC~\cite{corno2002ITC} and OpenCores~\cite{albrecht2005opencore}. For combinational AIGs, we use ForgeEDA~\cite{shi2025forgeeda}. For Sequential AIGs, we follow DeepSeq2~\cite{khan2025deepseq2} and extract sub-circuits from ITC~\cite{corno2002ITC}, OpenCores~\cite{albrecht2005opencore} and ISCAS'89~\cite{brglez1989ISCAS}. For PM netlist, we follow prior work~\cite{shi2025deepcell} and extract sub-circuits from ForgeEDA~\cite{shi2025forgeeda} and DeepCircuitX~\cite{li2025deepcircuitx}. 
We summarize the statistics of the datasets used in both contrastive tasks (Table~\ref{tab:dataset_contrastive}) and predictive tasks (Table~\ref{tab:dataset_predictive}). Since the global function in Definition~\ref{def:global_func} is hard to acquire, we follow previous works~\cite{shi2023deepgate2,shi2025deepcell} and approximate it using random simulations.
\begin{table}[h]
\centering
\caption{Dataset for contrastive tasks with $\{min. , avg., max.\}$.}
\label{tab:dataset_contrastive}
\resizebox{0.95\linewidth}{!}{
\begin{tabular}{lccc}
\toprule
 & RTL & AIG & PM Netlist \\ \midrule
\#Graphs & 1138 & 67706 & 67728 \\
\#Nodes & \{ 9, 102.6, 1987\} & \{22, 162.2,   2127\} & \{16, 88.6,   1192\} \\
\#Edges & \{10, 278.3, 2645 \} & \{22, 176.0,   2275\} & \{20, 110.5,   1361\} \\
Depth & \{2,  8.9, 27\} & \{6, 16.7, 29\} & \{2, 5.9, 12\} \\ \bottomrule
\end{tabular}
}
\vspace{-3pt}
\end{table}

\begin{table}[h]
\centering
\vspace{-5pt}
\caption{Dataset for predictive tasks with $\{min. , avg., max.\}$.}
\label{tab:dataset_predictive}
\resizebox{0.95\linewidth}{!}{
\begin{tabular}{lccc}
\toprule
 & Com. AIG & Seq. AIG & PM Netlist \\ \midrule
\#Graphs & 9800  &10007  & 83042 \\
\#Nodes & \{10, 2324.9, 45409\} & \{23, 236.5, 1281\} & \{12, 668.1, 4783\} \\
\#Edges & \{12, 3268.57, 68676\} & \{21, 260.5, 1915\} & \{11, 1113.9, 8914\} \\
Depth & \{4, 49.1, 2657\} & \{4, 21.0, 102\} & \{1, 16.0, 260\} \\ \bottomrule
\end{tabular}
}
\vspace{-10pt}
\end{table}

\paragraph{\textbf{Evaluation Metrics}}
In this work, we evaluate our model on two types of tasks: contrastive and predictive.
For the first type, the contrastive task, the goal is retrieval. Given a query circuit, the model must identify its functionally equivalent (positive) counterpart from a pool of $N$ candidate circuits. For our experiments, we set the pool size to $N=256$ for RTL and $N=1024$ for AIG and PM Netlists. We measure performance using the Recall@k (Rec@k) metric, reporting scores for $k \in \{1, 5, 10\}$.
For the second type, the predictive tasks, we assess the model's ability to determine node-level properties. We follow previous works~\cite{shi2023deepgate2,khan2025deepseq2,shi2025deepcell} and perform logic-1 probability prediction, similarity prediction and transition probability prediction.
We evaluate the accuracy of these predictions using Mean Absolute Error (MAE) and the $R^2$ score.

\vspace{-5pt}
\subsection{Contrastive Tasks}
As shown in Table~\ref{tab:contrastive_result}, \ourmethod\ consistently outperforms all baselines across the RTL, AIG, and PM Netlist modalities. Traditional message-passing GNNs (GCN, GIN, etc.) and specialized encoders (CircuitEncoder, FGNN2, DeepCell) achieve moderate performance, with Rec@1 scores generally ranging from 80\% to 88.7\%. However, Graph Transformer variants (GraphGPS, SGFormer, DIFFormer) show highly variable results: while competitive on RTL (up to 88.28\%), they suffer severe performance degradation on AIGs and PM Netlists (Rec@1 dropping as low as 15.4\% or below 50\%) and can face OOM errors, highlighting their difficulty in handling circuit graph hierarchy. In contrast, \ourmethod\ achieves superior and stable results, pushing Rec@1 to 94.45\% (RTL), 92.68\% (AIGs), and 90.81\% (PM Netlists), with Rec@10 consistently to 99.9\%, 99.5\% and 99.4\% respectively. These results, which outperforms the second-best method by a large margin on RTL, AIG, and PM Netlist modalities, demonstrates its strong generalization capacity.

\begin{table*}[h]
  \centering
  \caption{Comparison of predictive tasks on combinational and sequential AIGs.}
  \renewcommand{\arraystretch}{0.85}
  \begin{tabular}{lcccccccc}
    \toprule
    \multicolumn{1}{c}{\multirow{3}{*}{\textbf{Model}}} & \multicolumn{4}{c}{\textbf{Combinational AIG}} & \multicolumn{4}{c}{\textbf{Sequential AIG}} \\
    \cmidrule(lr){2-5} \cmidrule(lr){6-9}
    & \multicolumn{2}{c}{Logic-1 Probability} & \multicolumn{2}{c}{Similarity Prediction} & \multicolumn{2}{c}{Logic-1 Probability} & \multicolumn{2}{c}{Transition Probability} \\
    \cmidrule(lr){2-3} \cmidrule(lr){4-5} \cmidrule(lr){6-7} \cmidrule(lr){8-9}
    & R$^2$ & MAE & R$^2$ & MAE & R$^2$ & MAE & R$^2$ & MAE \\
    \midrule
    \multicolumn{9}{l}{\textit{Message Passing Neural Network}} \\ \midrule
    GCN & 0.644 & 0.152 & 0.271 & 0.090 & 0.868 & 0.064 & 0.744 & 0.024 \\
    GAT & 0.618 & 0.157 & 0.029 & 0.090 & 0.877 & 0.053 & 0.831 & 0.016 \\
    GIN & 0.669 & 0.144 & 0.445 & 0.076 & 0.962 & 0.035 & 0.790 & 0.023 \\
    GraphSAGE & 0.675 & 0.143 & 0.438 & 0.078 & 0.927 & 0.048 & 0.867 & 0.017 \\
    DeepGate2 & 0.983 & 0.028 & 0.502 & 0.069 & - & - & - & - \\
    PolarGate & 0.493 & 0.192 & 0.021 & 0.113 & - & - & - & - \\
    MGVGA & 0.666 & 0.145 & 0.418 & 0.077 & - & - & - & - \\
    DeepSeq2 & - & - & - & - & 0.979 & 0.025 & 0.908 & 0.014 \\ 
    \midrule
    \multicolumn{9}{l}{\textit{Graph Transformer}} \\ \midrule
    GraphGPS & OOM & OOM & OOM & OOM & 0.971 & 0.026 & 0.901 & 0.012 \\
    SGFormer & 0.516 & 0.175 & -0.072 & 0.117 & 0.878 & 0.056 & 0.596 & 0.026 \\
    DIFFormer & OOM & OOM & OOM & OOM & 0.701 & 0.097 & 0.416 & 0.034 \\
    DeepGate4 & 0.984 & 0.027 & 0.464 & 0.078 & - & - & - & - \\
     \midrule
    \textbf{\ourmethod} & \textbf{0.989} & \textbf{0.015} & \textbf{0.633} & \textbf{0.055} & \textbf{0.997} & \textbf{0.009} & \textbf{0.976} & \textbf{0.005} \\ \bottomrule
  \end{tabular}
  \label{tab:predictive_aig}
  \vspace{-5pt}
\end{table*}

\begin{table*}[h]
\caption{Comparison of predictive task on PM netlists.}
\label{tab:pm}

\begin{tabular}{@{}l ccccc ccc c@{}}
\toprule
\multirow{2}{*}{Metric} & \multicolumn{5}{c}{Message Passing Neural Network} & \multicolumn{3}{c}{Graph Transformer} & \multirow{2}{*}{\textbf{\ourmethod}} \\ 
\cmidrule(lr){2-6} \cmidrule(lr){7-9} 
 & GCN & GraphSAGE & GAT & GIN & DeepCell & GraphGPS & SGformer & DIFFormer & \\ \midrule
R$^2$ & 0.718 & 0.946 & 0.902 & 0.734 & 0.942 & 0.846 & 0.918 & 0.696 & \textbf{0.994} \\
MAE & 0.112 & 0.048 & 0.059 & 0.102 & 0.053 & 0.083 & 0.056 & 0.141 & \textbf{0.013} \\ \bottomrule
\end{tabular}
\vspace{-5pt}
\end{table*}

\vspace{-7pt}

\subsection{Predictive Tasks}

As shown in Table~\ref{tab:predictive_aig} and Table~\ref{tab:pm}, \ourmethod\ demonstrates a clear and consistent superiority across all predictive tasks by achieving minimal MAE and best $R^2$, validating its strong generalization. On Combinational AIGs (Table~\ref{tab:predictive_aig}), where traditional message-passing baselines and Graph Transformers struggle, \ourmethod\ delivers outstanding performance. \ourmethod\ achieves an MAE of 0.015 for logic-1 probability prediction and significantly outperforms all baselines in similarity prediction with an MAE of 0.055. 
The performance gap is further emphasized on Sequential AIGs, whose temporal dependencies pose additional challenges. Here, \ourmethod\ achieves substantial gains, reaching a minimal MAE of 0.009 for logic-1 probability prediction and an MAE of 0.005 for transition probability prediction, demonstrating superior precision. 
Finally, on PM Netlists (Table~\ref{tab:pm}), \ourmethod\ again surpasses all baselines, achieving a minimal MAE of 0.013. This consistent ability to minimize error across modalities, in contrast to the varied performance of other models, emphasizes the adaptability of our approach. 
Overall, by consistently delivering the minimal MAE scores, \ourmethod\ demonstrates its robust capacity to generalize across diverse combinational and sequential circuit graph domains.

\begin{figure}[htb]
    \centering
    \includegraphics[width=0.99\linewidth]{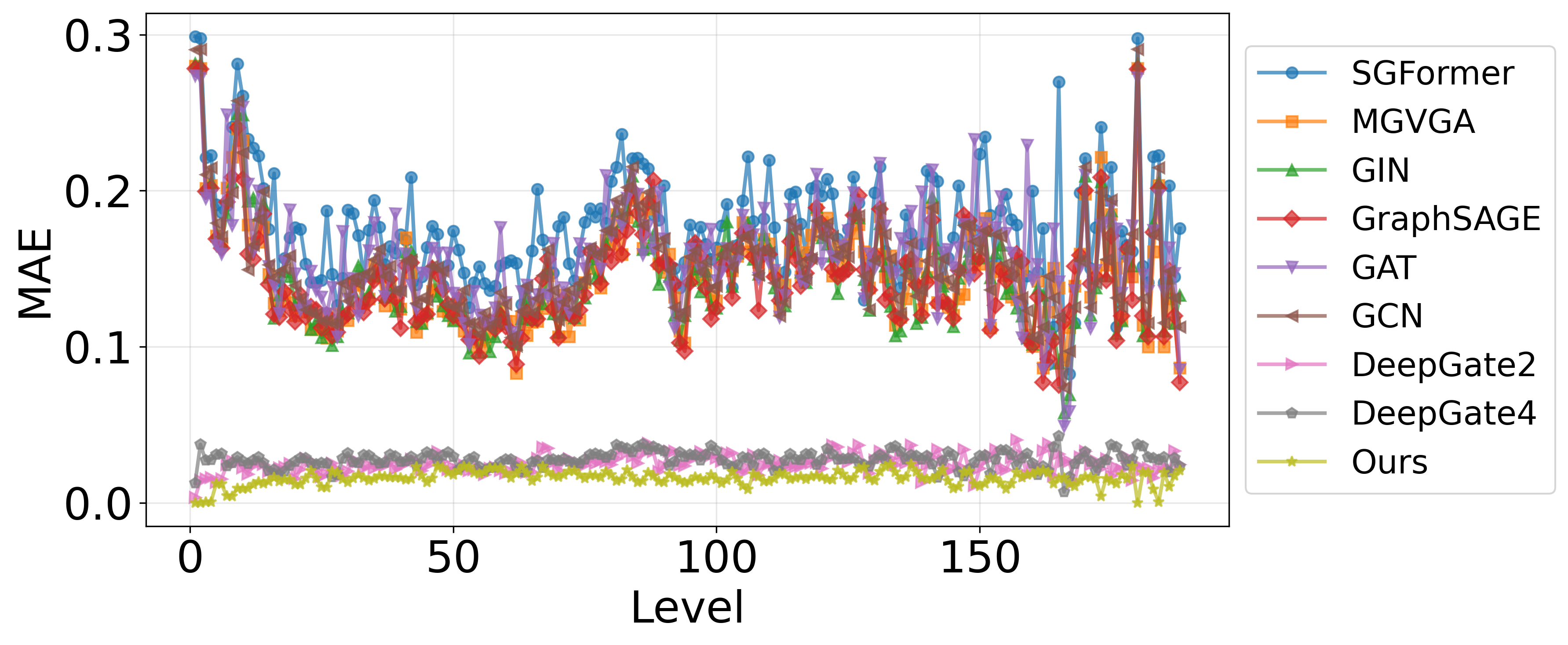}
    \\
    \vspace{-10pt}
    \includegraphics[width=0.99\linewidth]{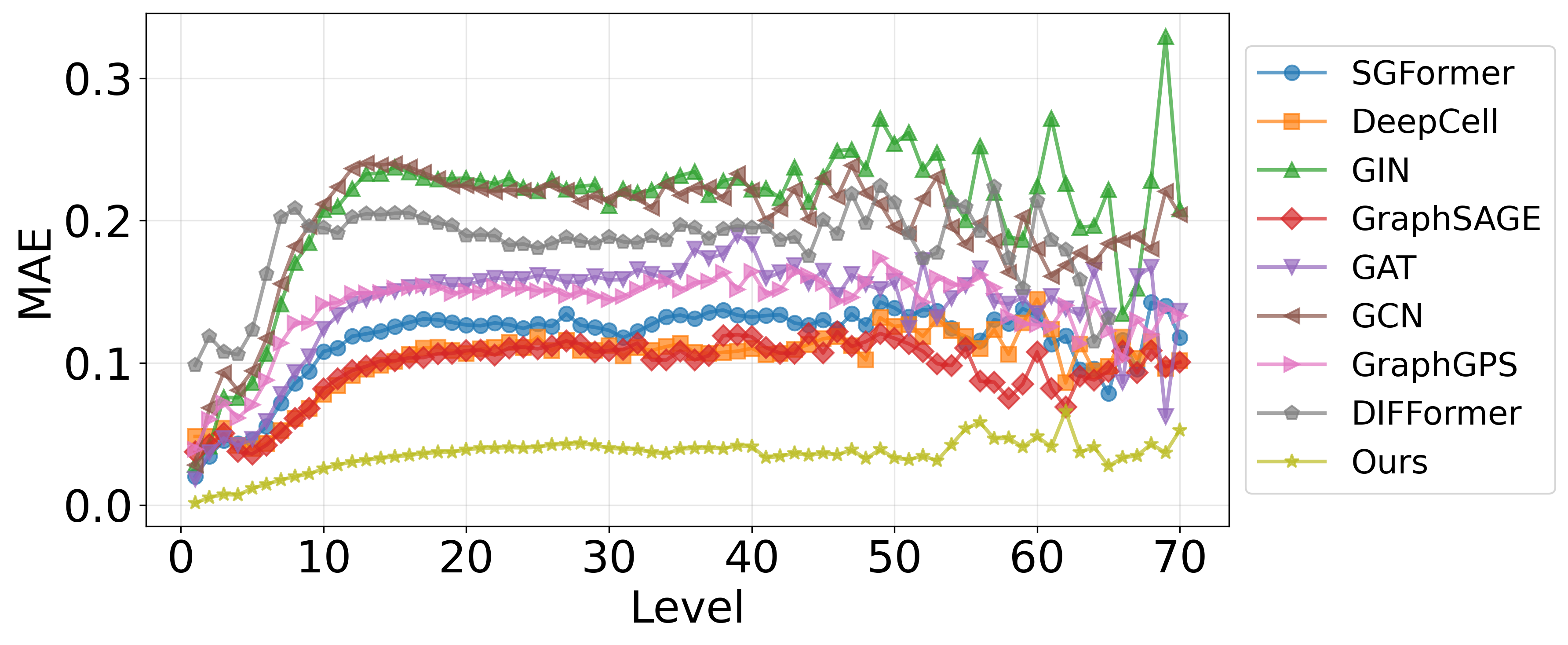}
    \vspace{-5pt}
    \caption{Error Accumulation Analysis. We illustrate the MAE of logic-1 probability prediction across different logic levels. Levels containing fewer than 10 nodes are excluded for better visualization. \textbf{Top:} Results on AIGs. \textbf{Bottom:} Results on PM netlists.}
    \vspace{-20pt}
    \label{fig:error_accumulation}
\end{figure}

\vspace{-5pt}
\paragraph{\textbf{Error Accumulation}} The iterative inference procedure in Algorithm~\ref{algo:inference} raises potential concerns about error accumulation. We investigate this by analyzing performance stability across different logic levels. As shown in Figure~\ref{fig:error_accumulation}, we plot the MAE for the logic-1 probability prediction task, grouping nodes by their logic level. The results demonstrate that \ourmethod\ maintains a stable and outstanding performance across all levels. This indicates the robustness of \ourmethod\ to potential error accumulation and generalizes effectively to circuit graphs of varying depths.

\begin{table}[]
\setlength{\tabcolsep}{2pt}
\caption{Ablation study on function shift learning (FSL).}
\resizebox{\linewidth}{!}{
\begin{tabular}{clcccccccc}
\toprule
\multicolumn{2}{c}{\multirow{2}{*}{Task}} & \multicolumn{2}{c}{GraphSAGE} & \multicolumn{2}{c}{DeepCell} & \multicolumn{2}{c}{DeepGate4} & \multicolumn{2}{c}{TRACE} \\ \cmidrule(l){3-10} 
\multicolumn{2}{c}{} & w/o & w. & w/o & w. & w/o & w. & w/o & w. \\ \midrule
\multirow{2}{*}{AIG} & Prob. & 0.143 & 0.043 & - & - & 0.027 & 0.020 & 0.024 & \textbf{0.015} \\
 & Sim. & 0.078 & 0.071 & - & - & 0.078 & 0.076 & 0.066 & \textbf{0.055} \\ \midrule
PM Netlist & Prob. & 0.048 & 0.046 & 0.053 & 0.047 & - & - & 0.036 & \textbf{0.013} \\ \bottomrule
\end{tabular}
}
\label{tab:ablation}
  \vspace{-10pt}
\end{table}

  

\vspace{-7pt}
\subsection{Ablation Study}


We conducted an ablation study to quantify the contribution of our proposed Function Shift Learning (FSL) component. The results in Table~\ref{tab:ablation}, compare our model (TRACE), second-based synchronous MPNNs (GraphSAGE) and asynchronous MPNNs (DeepGate4, DeepCell) with (w.) and without (w/o) the FSL component.

The findings demonstrate that FSL is crucial for achieving high performance. Our model, TRACE with FSL, consistently outperforms its ablated counterpart across all tasks. The impact is particularly significant on the PM Netlist probability prediction, where FSL reduces the MAE from 0.036 to 0.013, a relative error reduction of 63.9\%. For the AIG tasks, FSL also yields significant gains, reducing the probability prediction MAE by 37.5\% (from 0.024 to 0.015) and the similarity prediction MAE by 16.7\% (from 0.066 to 0.055).

While FSL provides a consistent performance boost for baseline MPNNs on AIGs, they gain only a slight boost on the PM Netlists modality. This suggests that standard MPNNs struggle to leverage FSL on circuit graphs where nodes represent more complex functions, highlighting their limitation in capturing both the global function the function shift—a gap TRACE is designed to address.

\vspace{-5pt}
\section{Conclusion}
In this work, we introduced TRACE, a new paradigm for learning on circuit graphs that addresses the architectural limitations of conventional MPNNs and Transformers. By employing a novel Hierarchical Transformer and a function shift learning objective, TRACE directly models the position-aware, hierarchical nature of computation. Our extensive experiments on electronic circuits demonstrate that TRACE substantially outperforms all prior architectures, establishing a new state of the art. This work provides a proof of principle for a more architecturally sound approach to learning on circuit graphs, offering a powerful framework with potential applications across various domains.

\balance


\newpage

\balance
\bibliographystyle{ACM-Reference-Format}
\bibliography{reference}

@article{chen2024large,
  title={Large circuit models: opportunities and challenges},
  author={Chen, Lei and Chen, Yiqi and Chu, Zhufei and Fang, Wenji and Ho, Tsung-Yi and Huang, Ru and Huang, Yu and Khan, Sadaf and Li, Min and Li, Xingquan and others},
  journal={Science China Information Sciences},
  volume={67},
  number={10},
  pages={200402},
  year={2024},
  publisher={Springer}
}

@ARTICLE{xie2022placement,
  author={Xie, Zhiyao and Liang, Rongjian and Xu, Xiaoqing and Hu, Jiang and Chang, Chen-Chia and Pan, Jingyu and Chen, Yiran},
  journal={IEEE Transactions on Computer-Aided Design of Integrated Circuits and Systems}, 
  title={Preplacement Net Length and Timing Estimation by Customized Graph Neural Network}, 
  year={2022},
  volume={41},
  number={11},
  pages={4667-4680},
  doi={10.1109/TCAD.2022.3149977}}

@INPROCEEDINGS{zhang2020grannite,
  author={Zhang, Yanqing and Ren, Haoxing and Khailany, Brucek},
  booktitle={2020 57th ACM/IEEE Design Automation Conference (DAC)}, 
  title={GRANNITE: Graph Neural Network Inference for Transferable Power Estimation}, 
  year={2020},
  pages={1-6},
  doi={10.1109/DAC18072.2020.9218643}}

@inproceedings{mendis2019ithemal,
title = "IThemal: Accurate, Portable and Fast Basic Block Throughput Estimation using Deep Neural Networks",
author = "Charith Mendis and Alex Renda and Saman Amarasinghe and Michael Carbin",
year = "2019",
language = "English (US)",
series = "36th International Conference on Machine Learning, ICML 2019",
publisher = "International Machine Learning Society (IMLS)",
pages = "7908--7918",
booktitle = "36th International Conference on Machine Learning, ICML 2019",
}

@INPROCEEDINGS{li2023deepsat,
  author={Li, Min and Shi, Zhengyuan and Lai, Qiuxia and Khan, Sadaf and Cai, Shaowei and Xu, Qiang},
  booktitle={2023 60th ACM/IEEE Design Automation Conference (DAC)}, 
  title={On EDA-Driven Learning for SAT Solving}, 
  year={2023},
  volume={},
  number={},
  pages={1-6},
  doi={10.1109/DAC56929.2023.10248001}}

@INPROCEEDINGS{zhang2021simulate,
  author={Zhang, He-Teng and Jiang, Jie-Hong R. and Amarú, Luca and Mishchenko, Alan and Brayton, Robert},
  booktitle={2021 58th ACM/IEEE Design Automation Conference (DAC)}, 
  title={Deep Integration of Circuit Simulator and SAT Solver}, 
  year={2021},
  volume={},
  number={},
  pages={877-882},
  doi={10.1109/DAC18074.2021.9586331}}

@article{Selsam2018LearningAS,
  title={Learning a SAT Solver from Single-Bit Supervision},
  author={Daniel Selsam and Matthew Lamm and Benedikt B{\"u}nz and Percy Liang and Leonardo Mendonça de Moura and David L. Dill},
  journal={ArXiv},
  year={2018},
  volume={abs/1802.03685},
  url={https://api.semanticscholar.org/CorpusID:3632319}
}

@INPROCEEDINGS{zuo2023mul,
  author={Zuo, Dongsheng and Ouyang, Yikang and Ma, Yuzhe},
  booktitle={2023 60th ACM/IEEE Design Automation Conference (DAC)}, 
  title={RL-MUL: Multiplier Design Optimization with Deep Reinforcement Learning}, 
  year={2023},
  volume={},
  number={},
  pages={1-6},
  doi={10.1109/DAC56929.2023.10247941}}

@article{Mirhoseini2021AGP,
  title={A graph placement methodology for fast chip design},
  author={Azalia Mirhoseini and Anna Goldie and Mustafa Yazgan and Joe Wenjie Jiang and Ebrahim M. Songhori and Shen Wang and Young-Joon Lee and Eric Johnson and Omkar Pathak and Azade Nazi and Jiwoo Pak and Andy Tong and Kavya Srinivasa and Will Hang and Emre Tuncer and Quoc V. Le and James Laudon and Richard Ho and Roger Carpenter and Jeff Dean},
  journal={Nature},
  year={2021},
  volume={594},
  pages={207 - 212},
  url={https://api.semanticscholar.org/CorpusID:235395490}
}

@inproceedings{fang2025circuitfusion,
 author = {Fang, Wenji and Liu, Shang and Wang, Jing and Xie, Zhiyao},
 booktitle = {International Conference on Representation Learning},
 editor = {Y. Yue and A. Garg and N. Peng and F. Sha and R. Yu},
 pages = {51523--51545},
 title = {CircuitFusion: Multimodal Circuit Representation Learning for Agile Chip Design},
 url = {https://proceedings.iclr.cc/paper_files/paper/2025/file/7ffb43adf37b3eeaba559098bc084cc6-Paper-Conference.pdf},
 volume = {2025},
 year = {2025}
}

@inproceedings{fang2025circuitencoder,
  title={A self-supervised, pre-trained, and cross-stage-aligned circuit encoder provides a foundation for various design tasks},
  author={Fang, Wenji and Liu, Shang and Zhang, Hongce and Xie, Zhiyao},
  booktitle={Proceedings of the 30th Asia and South Pacific Design Automation Conference},
  pages={505--512},
  year={2025}
}

@article{corno2002ITC,
  title={RT-level ITC'99 benchmarks and first ATPG results},
  author={Corno, Fulvio and Reorda, Matteo Sonza and Squillero, Giovanni},
  journal={IEEE Design \& Test of computers},
  volume={17},
  number={3},
  pages={44--53},
  year={2002},
  publisher={IEEE}
}

@inproceedings{albrecht2005opencore,
  title={IWLS 2005 benchmarks},
  author={Albrecht, Christoph},
  booktitle={International Workshop for Logic Synthesis (IWLS)},
  volume={9},
  year={2005}
}

@article{shi2025forgeeda,
  title={ForgeEDA: A Comprehensive Multimodal Dataset for Advancing EDA},
  author={Shi, Zhengyuan and Li, Zeju and Ma, Chengyu and Zhou, Yunhao and Zheng, Ziyang and Liu, Jiawei and Pan, Hongyang and Zhou, Lingfeng and Li, Kezhi and Zhu, Jiaying and others},
  journal={arXiv preprint arXiv:2505.02016},
  year={2025}
}

@inproceedings{khan2025deepseq2,
  title={Deepseq2: Enhanced sequential circuit learning with disentangled representations},
  author={Khan, Sadaf and Shi, Zhengyuan and Zheng, Ziyang and Li, Min and Xu, Qiang},
  booktitle={Proceedings of the 30th Asia and South Pacific Design Automation Conference},
  pages={498--504},
  year={2025}
}

@techreport{brglez1989ISCAS,
  title={Notes on the ISCAS'89 Benchmark Circuits},
  author={Brglez, Franc and Bryan, David and Kozminski, Krzysztof},
  year={1989},
  institution={Technical report, MCNC, 1989. Online http://www. cbl. ncsu. edu/CBL Docs~…}
}

@article{li2025deepcircuitx,
  title={Deepcircuitx: A comprehensive repository-level dataset for rtl code understanding, generation, and ppa analysis},
  author={Li, Zeju and Xu, Changran and Shi, Zhengyuan and Peng, Zedong and Liu, Yi and Zhou, Yunhao and Zhou, Lingfeng and Ma, Chengyu and Zhong, Jianyuan and Wang, Xi and others},
  journal={arXiv preprint arXiv:2502.18297},
  year={2025}
}

@article{shi2025deepcell,
  title={Deepcell: Multiview representation learning for post-mapping netlists},
  author={Shi, Zhengyuan and Ma, Chengyu and Zheng, Ziyang and Zhou, Lingfeng and Pan, Hongyang and Jiang, Wentao and Yang, Fan and Yang, Xiaoyan and Chu, Zhufei and Xu, Qiang},
  journal={arXiv preprint arXiv:2502.06816},
  year={2025}
}

@article{rampavsek2022graphgps,
  title={Recipe for a general, powerful, scalable graph transformer},
  author={Ramp{\'a}{\v{s}}ek, Ladislav and Galkin, Michael and Dwivedi, Vijay Prakash and Luu, Anh Tuan and Wolf, Guy and Beaini, Dominique},
  journal={Advances in Neural Information Processing Systems},
  volume={35},
  pages={14501--14515},
  year={2022}
}

@article{zheng2025deepgate4,
  title={DeepGate4: Efficient and Effective Representation Learning for Circuit Design at Scale},
  author={Zheng, Ziyang and Huang, Shan and Zhong, Jianyuan and Shi, Zhengyuan and Dai, Guohao and Xu, Ningyi and Xu, Qiang},
  journal={arXiv preprint arXiv:2502.01681},
  year={2025}
}

@article{wang2024fgnn2,
  title={Fgnn2: A powerful pre-training framework for learning the logic functionality of circuits},
  author={Wang, Ziyi and Bai, Chen and He, Zhuolun and Zhang, Guangliang and Xu, Qiang and Ho, Tsung-Yi and Huang, Yu and Yu, Bei},
  journal={IEEE Transactions on Computer-Aided Design of Integrated Circuits and Systems},
  year={2024},
  publisher={IEEE}
}

@inproceedings{wu2023gamora,
  title={Gamora: Graph learning based symbolic reasoning for large-scale boolean networks},
  author={Wu, Nan and Li, Yingjie and Hao, Cong and Dai, Steve and Yu, Cunxi and Xie, Yuan},
  booktitle={2023 60th ACM/IEEE Design Automation Conference (DAC)},
  pages={1--6},
  year={2023},
  organization={IEEE}
}

@inproceedings{PolarGate,
  author={Liu, Jiawei and Zhai, Jianwang and Zhao, Mingyu and Lin, Zhe and Yu, Bei and Shi, Chuan},
  booktitle={2024 IEEE/ACM International Conference on Computer-Aided Design (ICCAD)}, 
  title={PolarGate: Breaking the Functionality Representation Bottleneck of And-Inverter Graph Neural Network}, 
  year={2024}
}

@article{shi2024deepgate3,
  title={DeepGate3: Towards Scalable Circuit Representation Learning},
  author={Zhengyuan Shi and Ziyang Zheng and Sadaf Khan and Jianyuan Zhong and Min Li and Qiang Xu},
  journal={2024 ACM/IEEE International Conference On Computer Aided Design (ICCAD)},
  year={2024},
  pages={1-9},
  url={https://api.semanticscholar.org/CorpusID:271217918}
}

@inproceedings{shi2023deepgate2,
  title={Deepgate2: Functionality-aware circuit representation learning},
  author={Shi, Zhengyuan and Pan, Hongyang and Khan, Sadaf and Li, Min and Liu, Yi and Huang, Junhua and Zhen, Hui-Ling and Yuan, Mingxuan and Chu, Zhufei and Xu, Qiang},
  booktitle={2023 IEEE/ACM International Conference on Computer Aided Design (ICCAD)},
  pages={1--9},
  year={2023},
  organization={IEEE}
}

@inproceedings{wu2025MGVGA,
 author = {Wu, Haoyuan and Zheng, Haisheng and Pu, Yuan and Yu, Bei},
 booktitle = {International Conference on Representation Learning},
 editor = {Y. Yue and A. Garg and N. Peng and F. Sha and R. Yu},
 pages = {86310--86326},
 title = {Circuit Representation Learning with Masked Gate Modeling and Verilog-AIG Alignment},
 url = {https://proceedings.iclr.cc/paper_files/paper/2025/file/d6c06b4cab132aaade78d4d4d930b9c8-Paper-Conference.pdf},
 volume = {2025},
 year = {2025}
}

@INPROCEEDINGS{fang2025nettag,
  author={Fang, Wenji and Li, Wenkai and Liu, Shang and Lu, Yao and Zhang, Hongce and Xie, Zhiyao},
  booktitle={2025 62nd ACM/IEEE Design Automation Conference (DAC)}, 
  title={NetTAG: A Multimodal RTL-and-Layout-Aligned Netlist Foundation Model via Text-Attributed Graph}, 
  year={2025},
  volume={},
  number={},
  pages={1-7},
  doi={10.1109/DAC63849.2025.11133349}}

@inproceedings{gilmer2017neural,
  title={Neural message passing for quantum chemistry},
  author={Gilmer, Justin and Schoenholz, Samuel S and Riley, Patrick F and Vinyals, Oriol and Dahl, George E},
  booktitle={International conference on machine learning},
  pages={1263--1272},
  year={2017},
  organization={Pmlr}
}

@article{infonce,
  title={Representation learning with contrastive predictive coding},
  author={Oord, Aaron van den and Li, Yazhe and Vinyals, Oriol},
  journal={arXiv preprint arXiv:1807.03748},
  year={2018}
}

@inproceedings{li2022deepgate,
  title={Deepgate: Learning neural representations of logic gates},
  author={Li, Min and Khan, Sadaf and Shi, Zhengyuan and Wang, Naixing and Yu, Huang and Xu, Qiang},
  booktitle={Proceedings of the 59th ACM/IEEE Design Automation Conference},
  pages={667--672},
  year={2022}
}

@inproceedings{hoga,
  title={Less is more: Hop-wise graph attention for scalable and generalizable learning on circuits},
  author={Deng, Chenhui and Yue, Zichao and Yu, Cunxi and Sarar, Gokce and Carey, Ryan and Jain, Rajeev and Zhang, Zhiru},
  booktitle={Proceedings of the 61st ACM/IEEE Design Automation Conference},
  pages={1--6},
  year={2024}
}

@article{alon2020bottleneck,
  title={On the bottleneck of graph neural networks and its practical implications},
  author={Alon, Uri and Yahav, Eran},
  journal={arXiv preprint arXiv:2006.05205},
  year={2020}
}

@inproceedings{li2018deeper,
  title={Deeper insights into graph convolutional networks for semi-supervised learning},
  author={Li, Qimai and Han, Zhichao and Wu, Xiao-Ming},
  booktitle={AAAI},
  volume={32},
  number={1},
  year={2018}
}

@article{wu2023sgformer,
  title={Sgformer: Simplifying and empowering transformers for large-graph representations},
  author={Wu, Qitian and Zhao, Wentao and Yang, Chenxiao and Zhang, Hengrui and Nie, Fan and Jiang, Haitian and Bian, Yatao and Yan, Junchi},
  journal={Advances in Neural Information Processing Systems},
  volume={36},
  pages={64753--64773},
  year={2023}
}

@article{wu2023difformer,
  title={Difformer: Scalable (graph) transformers induced by energy constrained diffusion},
  author={Wu, Qitian and Yang, Chenxiao and Zhao, Wentao and He, Yixuan and Wipf, David and Yan, Junchi},
  journal={arXiv preprint arXiv:2301.09474},
  year={2023}
}

@article{gin,
  title={How powerful are graph neural networks?},
  author={Xu, Keyulu and Hu, Weihua and Leskovec, Jure and Jegelka, Stefanie},
  journal={arXiv preprint arXiv:1810.00826},
  year={2018}
}

@article{graphsage,
  title={Inductive representation learning on large graphs},
  author={Hamilton, Will and Ying, Zhitao and Leskovec, Jure},
  journal={Advances in neural information processing systems},
  volume={30},
  year={2017}
}

@article{ArminBiere2018Btor2,
  title={Btor2 , BtorMC and Boolector3.0},
  author={ArminBiere and AinaNiemetz and MathiasPreiner and CliffordWolf},
  journal={Springer, Cham},
  year={2018},
}

\end{document}